\documentclass[conference]{IEEEtran}
\IEEEoverridecommandlockouts
\usepackage{cite}
\usepackage{amsmath,amssymb,amsfonts}
\usepackage{algorithmic}
\usepackage{graphicx}
\usepackage{textcomp}
\usepackage{subcaption}
\usepackage{xcolor}
\def\BibTeX{{\rm B\kern-.05em{\sc i\kern-.025em b}\kern-.08em
    T\kern-.1667em\lower.7ex\hbox{E}\kern-.125emX}}
\begin{document}

\title{\LARGE \bf 
DP and QP Based Decision-making and Planning for Autonomous Vehicle
}


\author{\IEEEauthorblockN{1\textsuperscript{st} Zhicheng Zhang}
\IEEEauthorblockA{\textit{James Watt School of Engineering} \\
\textit{University of Glasgow}\\
Glasgow, United Kingdom \\
2943209z@student.gla.ac.uk}
}

\maketitle

\begin{abstract}
Autonomous driving technology is rapidly evolving and becoming a pivotal element of modern automation systems. Effective decision-making and planning are essential to ensuring autonomous vehicles operate safely and efficiently in complex environments. This paper introduces a decision-making and planning framework for autonomous vehicles, leveraging dynamic programming (DP) for global path planning and quadratic programming (QP) for local trajectory optimization. The proposed approach utilizes S-T graphs to achieve both dynamic and static obstacle avoidance. A comprehensive vehicle dynamics model supports the control system, enabling precise path tracking and obstacle handling. Simulation studies are conducted to evaluate the system’s performance in a variety of scenarios, including global path planning, static obstacle avoidance, and dynamic obstacle avoidance involving pedestrian interactions. The results confirm the effectiveness and robustness of the proposed decision-making and planning algorithms in navigating complex environments, demonstrating the feasibility of this approach for autonomous driving applications.
\end{abstract}


\section{Introduction}

\subsection{Motivation and Problem Statement}

In recent years, autonomous driving technology has made significant advancements, becoming a cornerstone of modern transportation automation. Unlike drones or smart robots, autonomous vehicles are responsible for performing various ground transportation tasks while maintaining passenger comfort and vehicle stability. To ensure the safety and comfort of passengers, it is crucial that autonomous vehicles maintain stability in all driving conditions. Therefore, a reliable decision-making and planning system is required to navigate effectively in challenging scenarios, such as avoiding dynamically appearing obstacles, handling sharp turns, and ensuring pedestrian safety on roadways. Additionally, unstable driving and extra mileage can lead to increased tire wear and passenger discomfort, emphasizing the need for planning algorithms that can generate efficient, stable, and collision-free paths.

Before executing decision-making and planning algorithms, it is essential to establish a comprehensive understanding of the driving environment using global perception techniques, such as those described in~\cite{wong2020testing}~\cite{10354351}. These technologies enable autonomous vehicles to perceive and interpret their surroundings in real-time, accurately identifying obstacles and their positions, which is crucial for making informed decisions. Once the perception layer is in place, the focus shifts to designing effective path-planning and control strategies that ensure the vehicle's safety and stability in dynamic environments.

Planning algorithms for autonomous vehicles are responsible for generating safe and optimized routes that avoid potential collisions while remaining stable. In this context, dynamic programming (DP) and quadratic programming (QP) algorithms are essential, as they enable decision-making that balances optimal path selection with vehicle stability~\cite{fan2018baidu}. The S-T graph representation provides a visual framework for trajectory planning, allowing the vehicle to adapt to both static and dynamic obstacles.

This paper presents a decision-making and planning framework for autonomous vehicles, integrating DP and QP algorithms to achieve safe and stable navigation. Global planning is handled through dynamic programming to generate optimal routes, while local trajectory optimization is performed using quadratic programming to ensure smooth and collision-free paths. The S-T graph approach is employed to facilitate real-time obstacle avoidance and path adjustments based on the perceived environment.

The main contributions of this work are summarized as follows:
\begin{itemize}

\item Decision-making and Planning Framework: This paper proposes an integrated decision-making and planning framework based on DP for global planning and QP for local trajectory optimization. The combined approach allows the autonomous vehicle to navigate complex environments safely and efficiently.

\item Dynamic and Static Obstacle Avoidance: The proposed framework leverages an S-T graph to represent the vehicle’s planned path and speed profile in real time, enabling effective avoidance of both dynamic and static obstacles.

\item Enhanced Navigation Capabilities: By incorporating DP and QP algorithms, this work enhances the vehicle’s ability to generate optimized routes and trajectories, improving overall navigation performance and operational safety.

\end{itemize}

\subsection{Related Works}

Various controllers have been developed to achieve effective control of autonomous vehicles. Stanford University’s Stanley autonomous vehicle utilized a Proportional-Integral (PI) controller, which was designed based on a linearized model of the vehicle’s dynamics relative to the path \cite{thrun2006stanley}. A Radial Basis Function Neural Network (RBFNN)-based adaptive PID controller was proposed for longitudinal control to achieve precise speed tracking \cite{nie2018longitudinal}. Moreover, fuzzy logic controllers were integrated with traditional PID controllers to enable parameter adaptation, improving control accuracy \cite{li2022autonomous}. However, these methods often face challenges in achieving precise longitudinal control, as throttle and brake are separate components, and a single PID controller may not be efficient or accurate enough in coordinating both systems.

For lateral control, alternative techniques such as sliding mode control (SMC) have been employed to counter parameter uncertainties and external disturbances, allowing better trajectory control \cite{gerdes1997vehicle}. Additionally, fuzzy PID controllers focusing on new self-adaptive fuzzy PID designs based on nonlinear multi-input multi-output (MIMO) structures have been explored \cite{khodayari2015modeling}~\cite{cui2024design}. However, PID control remains a simple feedback mechanism with manually tuned parameters, making it difficult to achieve real-time and precise responses \cite{10387140}. In contrast, Linear Quadratic Regulator (LQR) control is designed using a state-space model of the system, enabling it to leverage the system’s dynamic characteristics more effectively \cite{9902113}.

For decision-making and planning algorithms, commonly employed methods include the A* search algorithm and Dijkstra’s algorithm \cite{10029975}. A typical decision-making and planning module can be divided into three levels: route planning, behavioral decision-making, and motion planning. Zhao et al. identified route planning as the process of generating a global path by combining map information after receiving a specified travel destination, which serves as a reference for subsequent detailed path planning \cite{zhao2024improved}. The behavioral decision-making layer makes decisions based on perception data and, through motion planning, ultimately generates a trajectory that satisfies specific constraints. This paper adopts a combination of DP and QP, which is well-suited for multi-stage decision problems and efficiently handles scenarios with large state spaces and complex state transitions \cite{8628351}.

\section{System Overview}

The proposed framework is a comprehensive control system for autonomous vehicles that integrates both longitudinal and lateral control to enhance overall safety and performance. The system employs a dual-loop PID controller to precisely manage the throttle and brake operations, maintaining stable longitudinal dynamics and smooth acceleration or deceleration. For lateral control, the use of a DLQR enables enhanced steering control and quick responsiveness to changes in the driving environment. Additionally, the framework incorporates advanced decision-making and path-planning strategies using DP and QP algorithms. These algorithms enable the system to generate optimized speed and trajectory plans, ensuring efficient navigation and obstacle avoidance. By leveraging DP and QP, the system can effectively handle complex scenarios involving both static and dynamic obstacles. Furthermore, the MPC controller is integrated to facilitate accurate trajectory tracking based on the planned paths. This combination of control and planning methodologies ensures reliable and adaptive vehicle navigation across varied environments, prioritizing passenger safety and ride comfort.

\subsection{Vehicle Model}

In this paper, to simplify the progress of trajectory generation, a bicycle model is employed as shown in Fig. 1, where $\varphi$ represents the yaw angle, $\beta$ represents the sideslip angle at the center of mass, $R$ represents the sideslip angle at the center of mass, $\delta_f$ and $\delta_r$  are the steering angles of the rear and front wheels respectively. Based on the obtained angular relationships, the kinematic equations are:

\begin{figure}[h]
    \centering
    \includegraphics[width=0.7\linewidth]{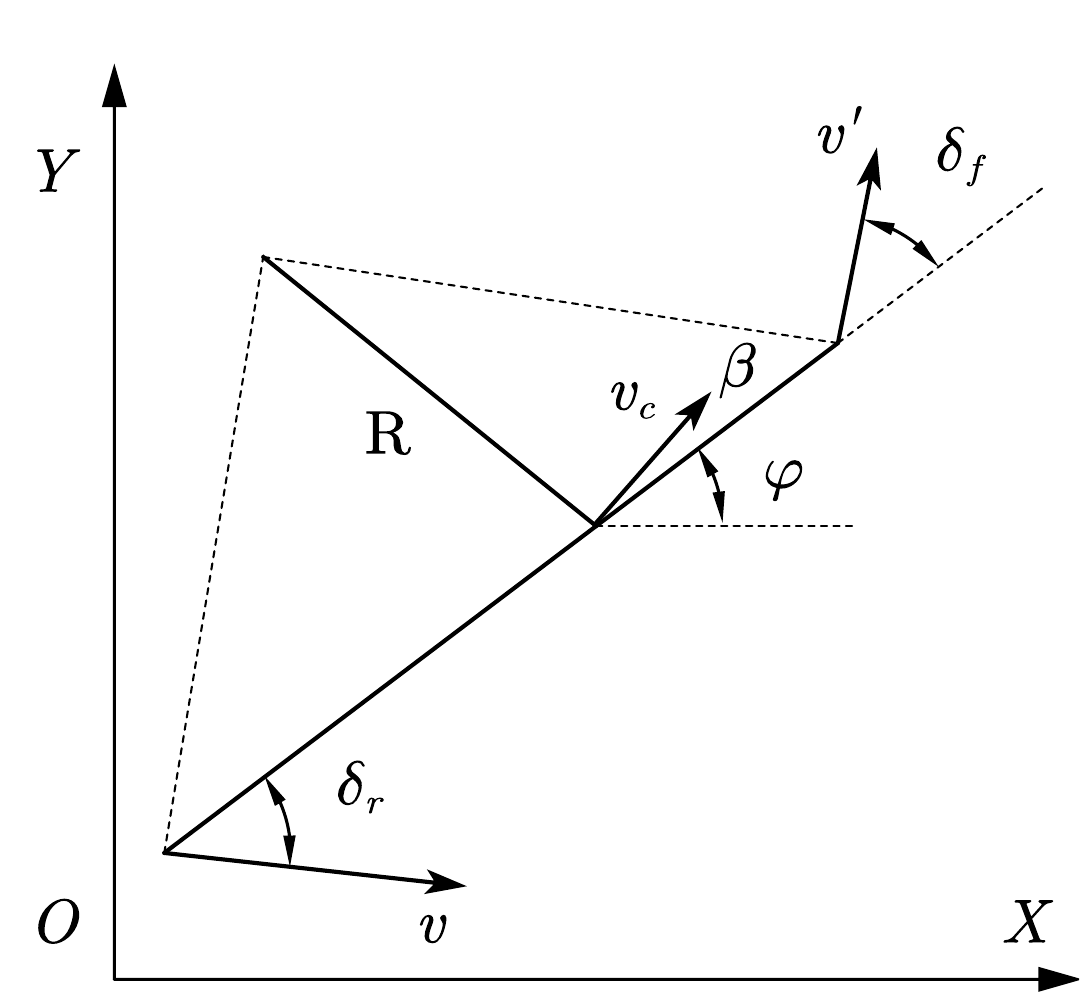}
    \, \, \, \, \, \caption{The bicycle model.}
    \label{fig.}
\end{figure}

\begin{equation}
\begin{split}
x &= v\cos (\theta + \beta) \\
y &= v\sin (\theta + \beta) \\
\varphi &= \frac{v(tan\delta_f +tan\delta_r)}{L}
\end{split}
\end{equation}

\begin{figure}[h]
    \centering
    \includegraphics[width=0.8\linewidth]{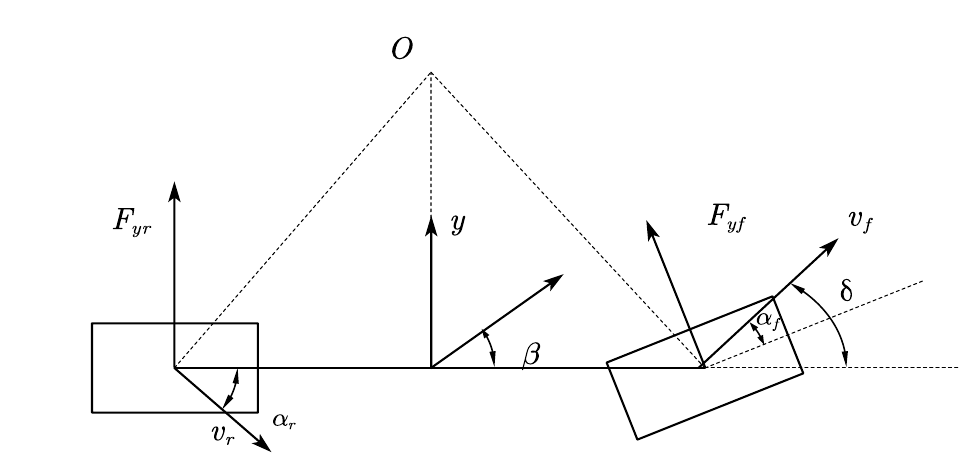}
    \, \, \, \, \, \caption{Side force diagram of the bicycle model.}
    \label{fig.}
\end{figure}

Where $L$ is the wheelbase of the vehicle. 

\subsection{DLQR Based Lateral Control}

Considering the impact of tire stiffness on generating lateral force, as shown in Fig. 2, $\alpha$ represents the tire slip angle and $F_y$ represents the slip force. The force and torque equations can be developed as:

\begin{equation}
\begin{split}
\sum F_y = m a_y = F_y f cos\delta + F_yr \\
\sum M = I \ddot{\varphi} = F_{yf} acos\delta - F_{yr}b
\end{split}
\end{equation}

Where $M$ represents the torque and $I$ is the moment of inertia. Assume $X = \left ( \begin{matrix}
 \dot{y}\\  \dot{\varphi}
\end{matrix} \right ) $, $u = \delta$, then $\dot{X} = AX+ Bu$. In which:

\begin{equation}
A = \left (  \begin{matrix}
 \frac{C_{\alpha f}+C_{ar}}{mv_x} & \frac{a C_{\alpha f} - b C_{\alpha r}}{mv_x}\\
\frac{a C_{\alpha f} - b C_{\alpha r}}{Iv_x}  & \frac{a^2 C_{\alpha f} + b^2 C_{\alpha r}}{Iv_x}
\end{matrix}\right ) 
\end{equation}

\begin{equation}
    B = (-\frac{C_{\alpha f}}{m}, \frac{aC_{\alpha f}}{I})^T
\end{equation}

Whereas $C_{\alpha f}$ and $C_{\alpha r}$ are the tire stiffness, $a$ and $b$ represent the front and rear wheelbase size. 

\begin{figure}[h]
    \centering
    \includegraphics[width=1\linewidth]{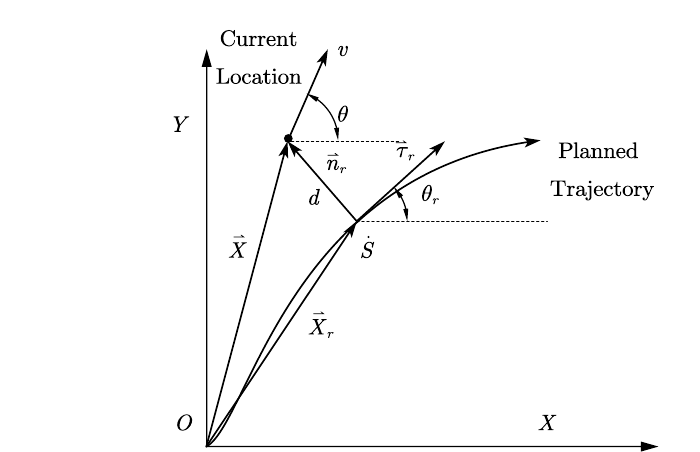}
    \, \, \, \, \, \caption{Error angle relationship.}
    \label{fig.3}
\end{figure}

In Fig. 3 the error between the planned trajectory and the actual trajectory is shown. In which $e_{rr}$ represents the error between the two trajectories. Then it can be derived as:

\begin{equation}
    \vec{e_{rr}} = \vec{x} - \vec{x_r} 
\end{equation}

Assume that $\dot{X} = \dot{e_{rr}}$, eqn(5) can be shifted as $ \dot{e_{rr}} = \bar{A} e_{rr} + \bar{B} u$. Therefore, the control objective becomes selecting an appropriate $u$ to make the planned trajectory close to the actual trajectory to minimize the trajectory error. To define the trajectory error, a cost function be derived as:

\begin{equation}
    J = e_{rr}^T Q e_{rr} + u^TRu
\end{equation}

Therefore an LQR can be deployed to minimize this cost function. As shown in Fig. 3, $d$ represents the lateral error, $\theta - \theta_r$ represents the heading angle error. $S$ is the distance of the vehicle.By integrating the angle relationship and applying the Fernet formula, the following equation can be derived:

\begin{equation}
    \begin{split}
\dot{d} = \left | \vec{v} \right | sin(\theta -\theta _r ) \\
\dot{S} = \frac{\left | \vec{v} \right | cos(\theta -\theta _r )}{1-kd}
    \end{split}
\end{equation}

In which $k$ represents the kappa of the curve. Then $e_{rr}$ can be represented as:

\begin{equation}
    \dot{e_{rr}} = A e_{rr} + Bu + C\dot{\theta_r}
\end{equation}

To minimize the cost function, a DLQR controlled is employed in this paper \cite{6940733}. The basic principle of DLQR can be presented as: 

\begin{equation}
    X_{k+1} = \bar{A}X_k + \bar{B}u_k
\end{equation}

Thus the cost function can be shifted as:

\begin{equation}
    J = \sum_{k=0}^{\infty } (x_{k}^{T}Qx_{k}+u_{k}^{T}Ru_{k})
\end{equation}

Using the boundary condition to minimize (9), discretization is typically performed using Euler's method. Using midpoint Euler rule for $x_k$ and the forward Euler rule for $u_k$:

\begin{equation}
    X(t+dt) = (I - \frac{Adt}{2})^{-1} (I + \frac{Adt}{2})X(t) +Bdtu(t)
\end{equation}

From the Riccati formula \cite{kumar2016algebraic}, $P_{k-1} = Q + A^{T}P_k(I+BR^{-1}B^{T}P_k)^{-1}A, P_n=Q$, then $u_k$ can be presented as:

\begin{equation}
    u = -(R+B^{T}P_{k+1}B)^{-1}B^{T}P_{k+1}AX_k=-KX_k
\end{equation}

With both $e_{rr}$ and $\dot{e_{rr}}$ both limited to $0$, $e_rr$ can be calculated through matrics $A, B, C$ and $K = (k_1, k_2 ,k_3, k_4)$.

\begin{figure}[h]
    \centering
    \includegraphics[width=0.8\linewidth]{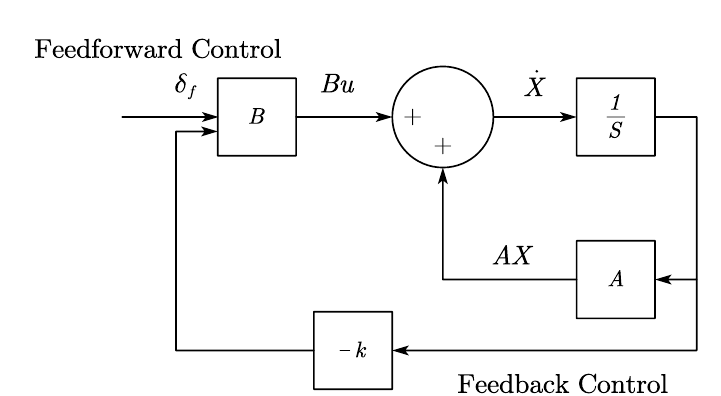}
    \, \, \, \, \, \caption{Feedforward control of DLQR controller.}
    \label{fig.}
\end{figure}

In Fig. 4, with feedforward control, $\dot{e_{rr}}$ can be further represented as:

\begin{equation}
    \dot{e_{rr}} = A e_{rr} + B(-ke_{rr} + \delta_f ) +C\dot{\theta_r}
\end{equation}

Then a suitable $\delta_f$ needs be considered to minimize the $e_{rr}$. Since $e_\varphi$ is not influenced by $\delta_f$ and $k$, simplify $e_\varphi$ using angle relationships:

\begin{equation}
    e_\varphi = -(\frac{b}{R} + \frac{a}{a+b}ma_y\frac{1}{C_r})
\end{equation}

Then $e_\varphi$ can be simplified as $e_\varphi  \doteq - \beta$.

\begin{figure}[h]
    \centering
    \includegraphics[width=0.8\linewidth]{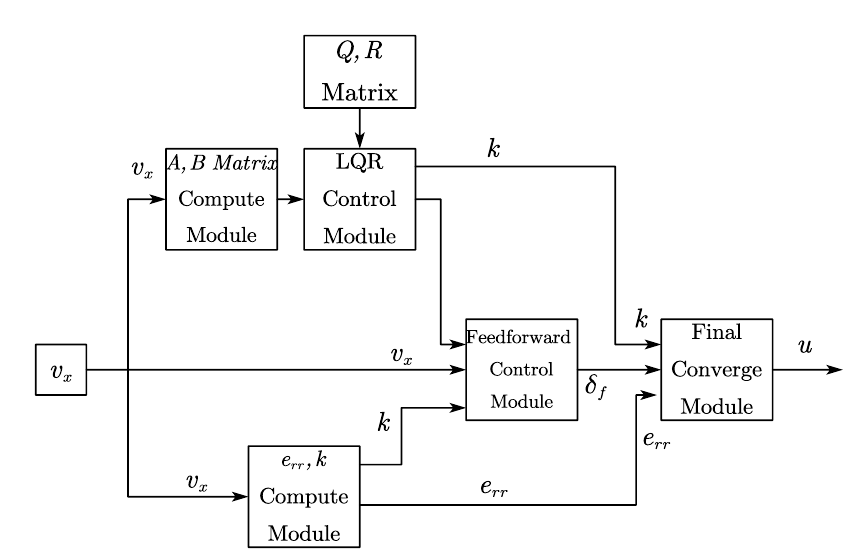}
    \, \, \, \, \, \caption{DLQR floatchat.}
    \label{fig.}
\end{figure}

The overall system's input consists of the vehicle's longitudinal velocity $v_x$ and its inherent properties, as illustrated in Fig. 5. The $A$ and $B$ matrix computation module determines the state matrix $A$ and input matrix $B$ based on $v_x$, which represent the system's dynamic model. The LQR control module then uses these matrices along with predefined weight matrices $Q$ and $R$ to calculate the feedback gain matrix $k$. This gain is transmitted to various modules, including those for feedforward control and error computation. The error computation module takes $v_x$ and $k$ to determine the system error. The feedforward control module computes the feedforward control input $\delta_f$ using $k$ and $v_x$ to pre-adjust the system's response. Finally, all control inputs $\delta_f$ and the computed error $e_{rr}$ are combined in the final convergence module, which integrates $k$ and $e_{rr}$  to produce the final control input $u$.

\subsection{Dual-loop PID Controller for Longitude Control}

Within the dual PID longitudinal control system, the Distance PID controller processes the distance error $e_{rr}$ derived from the lateral DLQR control, using it to compute the desired speed. This target speed is then fed into the Speed PID controller, which compares it with the vehicle's current speed to generate a control signal. This signal, along with the desired acceleration, is used to modulate throttle and brake inputs, ensuring the vehicle reaches and maintains the intended speed while following the planned trajectory. The tuning of the dual PID controllers primarily focuses on adjusting the proportional gain (KP), with the integral (KI) and derivative (KD) gains fine-tuned as necessary based on the throttle and brake response.

\subsection{MPC Based Trajectory Tracking}

This section will introduce how trajectory tracking is achieved using MPC. Firstly, to have a more comfortable ride, abrupt changes in these dynamics must be minimized \cite{liu2021accurate}. This is where the concept of $JERK$, the rate of change of acceleration, comes into play as an important metric for evaluating ride comfort, whcih can be defined as:

\begin{equation}
    JERK = \frac{da}{dt}
\end{equation}

Based on the definition of comfort, six related variables can be derived. A quintic polynomial is utilized to satisfy six specific constraints, making it ideal for trajectory control \cite{yue2019quintic}. This type of polynomial requires six boundary conditions to solve for its coefficients. These conditions typically include the initial and final values of position, velocity, and acceleration, providing a clear framework for defining the vehicle's motion from start to finish. The basic expression of a quintic polynomial is:

\begin{equation}
    f(t) = a_0 + a_1(t-t_0) + ... + a_6(t-t_0)^5
\end{equation}

The cost function based on MPC is obtained by solving the quintic polynomial as follows:

\begin{equation}
    J = \sum_{k=0}^{N} \left( Q_e \cdot \text{tracking\_error}_k + R_u \cdot \Delta u_k + R_{u\_rate} \cdot u_k \right)
\end{equation}

In which $Q_e$ is the weight for the tracking error, $R_u$ is the weight for the control input change and $R_{u_rate}$ represents the weight for the control input magnitude. With the implementation of the MPC controller, the process of generating a smooth curve and ensuring accurate trajectory tracking is finalized.

\subsection{DP and QP Based Obstacle Avoidance}

After completing the design of the autonomous driving control system, this paper incorporates trajectory planning capabilities based on DP and QP to generate $S-T$(State-Time Graph) figure into the controller, testing the obstacle avoidance scenarios for both static and dynamic obstacles.

For motion planning in autonomous driving, a lane-changing strategy is essential. One typical method is to develop an algorithm that explores all available lanes, but expanding the search space to multiple lanes increases the computational burden \cite{ajanovic2018search}. Moreover, it is critical to ensure trajectory stability to avoid abrupt transitions between planning cycles, as consistent on-road behavior is important to signal the autonomous vehicle’s intentions to other drivers \cite{werling2010optimal}. A parallel framework has been proposed to handle both passive and active lane changes. In this framework, all obstacles and environmental data are mapped onto lane-specific Frenet frames for each candidate lane. Lane-level strategies incorporate traffic rules, and an optimal path is generated for each lane using a dedicated optimizer. Finally, a cross-lane decision-making component selects the most suitable lane by assessing a cost function and adhering to safety protocols.

\begin{figure}[h]
    \centering
    \includegraphics[width=0.8\linewidth]{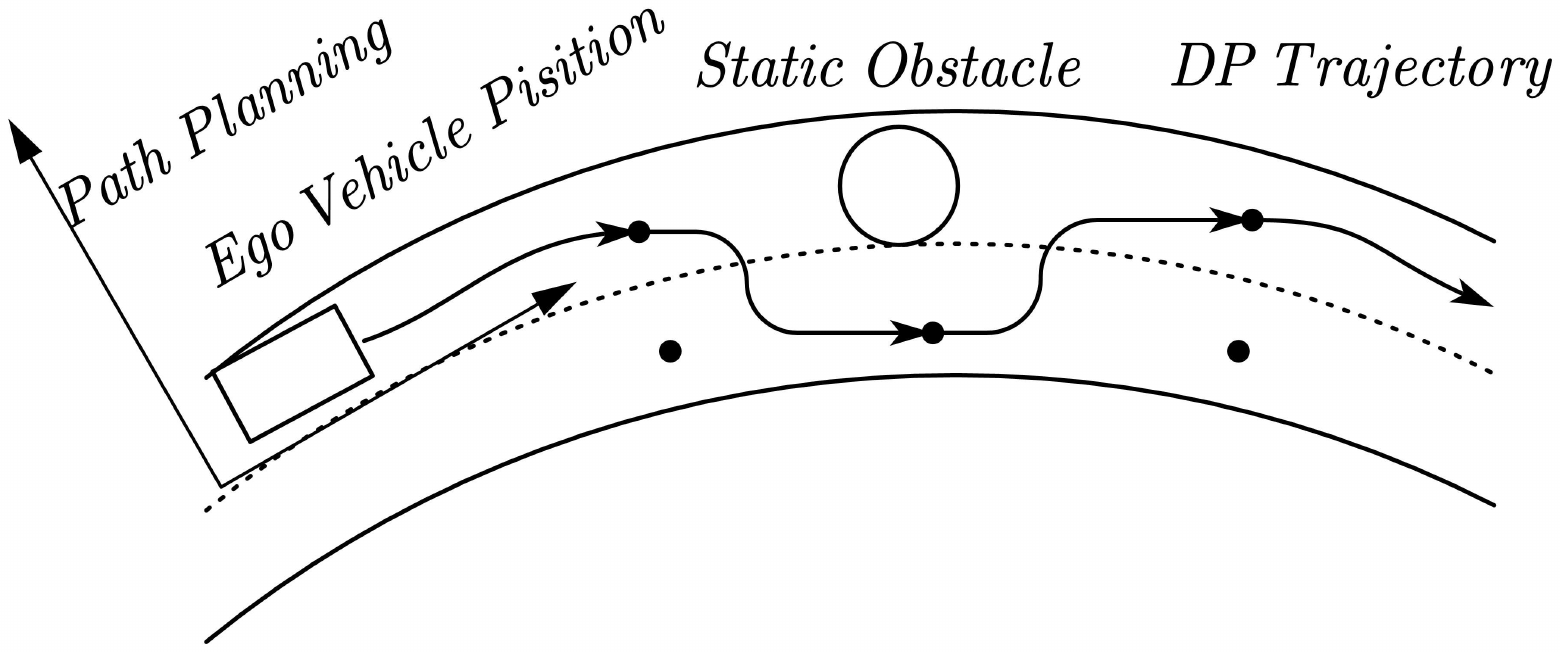}
    \, \, \, \, \, \caption{DP path planning.}
    \label{fig.}
\end{figure}

When dealing with static obstacles, as illustrated in Fig. 6, scattered points are evenly distributed around the planned path, constrained by the environment. For instance, when static obstacles are detected, the dynamic trajectory automatically analyzes and maps the obstacle coordinates into the Frenet coordinate system. This results in a convex space, providing two main directions for bypassing: either to the left or right, along with finer sub-directions for more detailed planning. Path optimization typically accounts for static obstacles by identifying an optimal route within the SLT framework through trajectory sampling or lattice-based searching. Once the path is generated, a corresponding speed profile is then established.

\begin{figure}[h]
    \centering
    \includegraphics[width=0.8\linewidth]{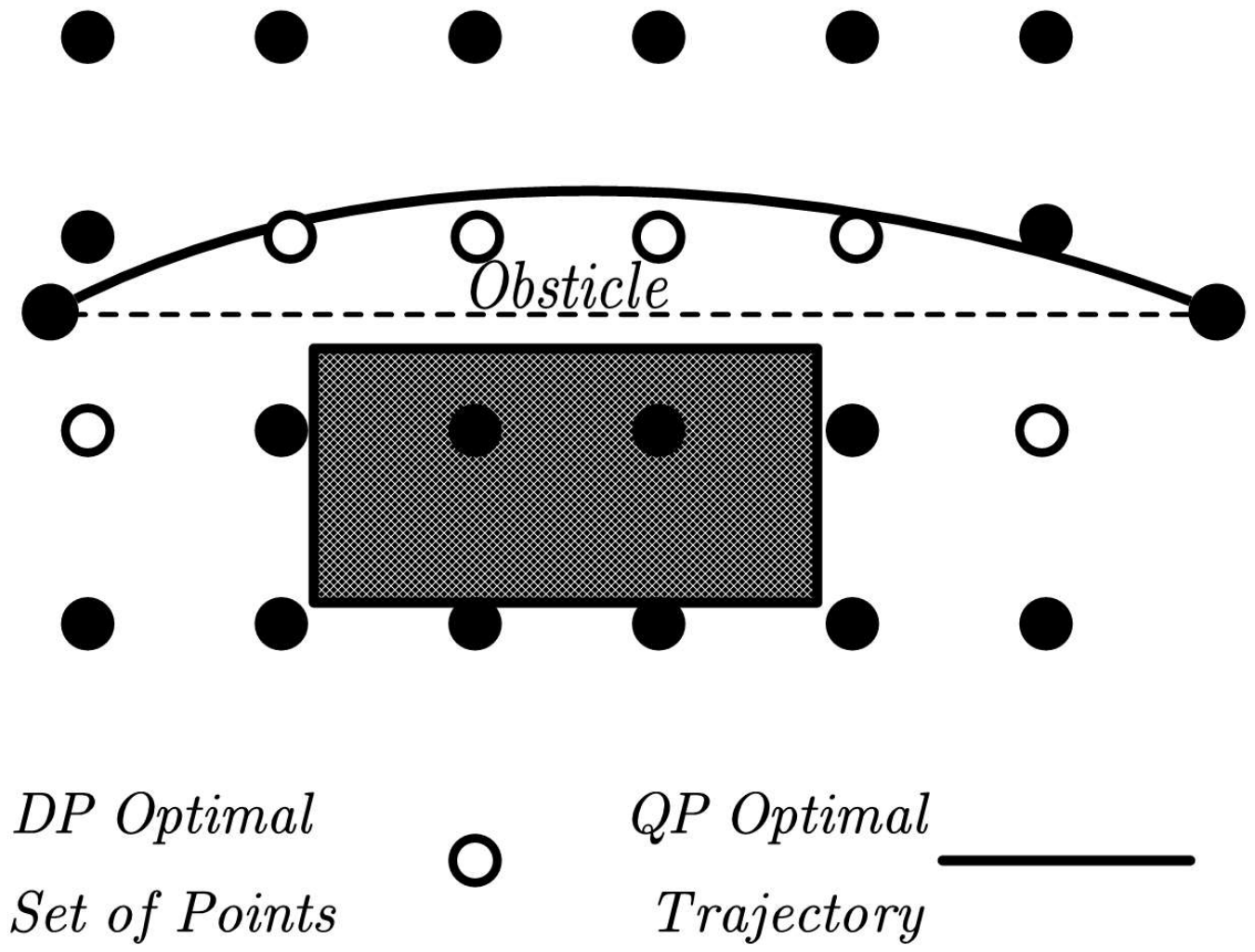}
    \, \, \, \, \, \caption{DP and QP path planning.}
    \label{fig.}
\end{figure}

Fig. 7. illustrates the trajectory planned using the QP approach. The cost function for the QP optimization process can be expressed as:

\begin{align}
    J_1 &= W_{ref}\cdot \sum_{i}l_{i}^{2} \notag \\
    J_2 &= W_{dl} \cdot \left (\sum_{i}l_{i}^{'} \right )^2 \notag \\
    J_3 &= W_{ddl} \cdot \left (\sum_{i}l_{i}^{''} \right )^2 \notag \\
    J_4 &= W_{dddl} \cdot \left [ \sum_{i}(l_{i+1}^{''}-l_{i}^{''})^2 \right ] \notag \\
    J_5 &= W_{mid} \cdot \left [ \sum_{i} \left ( l_i - \frac{l_{mini} + l_{maxi}}{2} \right )^2 \right ] \notag \\
    J &= J_1 + J_2 + J_3 + J_4 + J_5
\end{align}

In which $J_1$ refers to the cost of the reference line, $J_2, J_3, J_4$ represent the cost function for the first to third derivatives. $J_5$ is the center cost function of the convex space.

\begin{figure}[h]
    \centering
    \includegraphics[width=0.8\linewidth]{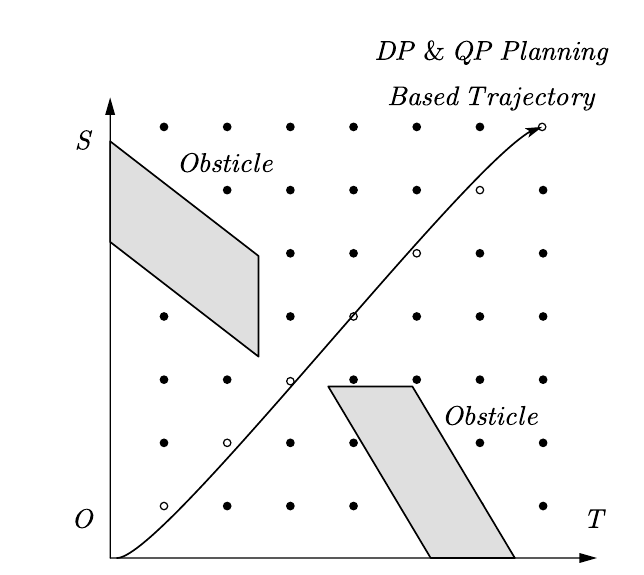}
    \, \, \, \, \, \caption{DP and QP path planning.}
    \label{fig.}
\end{figure}

Fig. 8 illustrates the $S-T$ Graph, a visual model that depicts the relationship between states and time in dynamic systems. It is particularly effective for addressing path planning and dynamic obstacle avoidance challenges \cite{felzenszwalb2010dynamicrefs}. In this graph, each node corresponds to the system's state at a given moment, including parameters like position and velocity. The edges signify the transitions between states, which can be affected by dynamic constraints and the characteristics of the system. By analyzing the generated $S-T$ graph, the obstacle avoidance performance of the autonomous vehicle can be evaluated. The next chapter will provide a detailed discussion of the conclusions.

\section{Simulation Results}

\subsection{Simulation Setup}\label{AA}

\begin{figure}[h]
    \centering
    \includegraphics[width=0.8\linewidth]{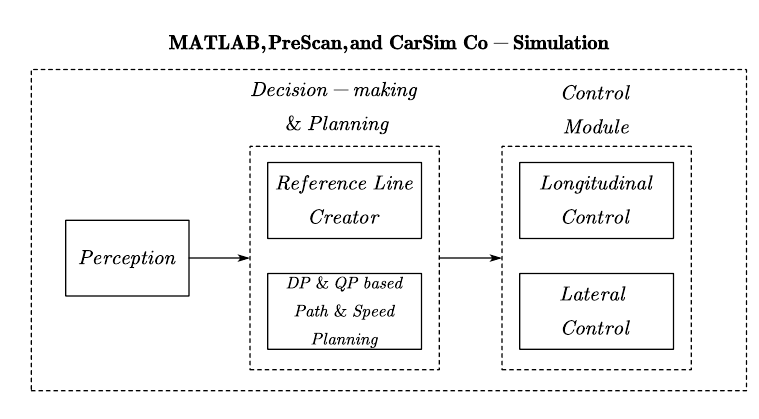}
    \, \, \, \, \, \caption{Co-simulation with CarSim and PreScan.}
    \label{fig.}
\end{figure}

This paper adopts a multi-software co-simulation strategy to model the autonomous driving system. Shown in Fig.8, initially, CarSim will be employed as a widely recognized software for vehicle dynamics simulation within both the automotive industry and academic research. It offers accurate vehicle models, a broad range of application scenarios, and compatibility with other software tools. Then, PreScan will be used to create the testing scenarios and set up the sensors. Lastly the controller and co-simulation will be deployed in MATLAB.

\subsection{Safety of Obstacle Avoidance}\label{AA}

The tests of this paper focus solely on the vehicle's trajectory tracking and obstacle avoidance, excluding considerations related to yielding or traffic safety. To this end, two distinct scenarios will be established. The first scenario will feature no obstacles, allowing for an assessment of the vehicle's tracking performance. In the second scenario, a pedestrian will cross laterally, and an $S-T$ diagram will be generated to analyze the vehicle's yielding and avoidance actions.

\begin{figure}[h]
    \centering
    \includegraphics[width=0.8\linewidth]{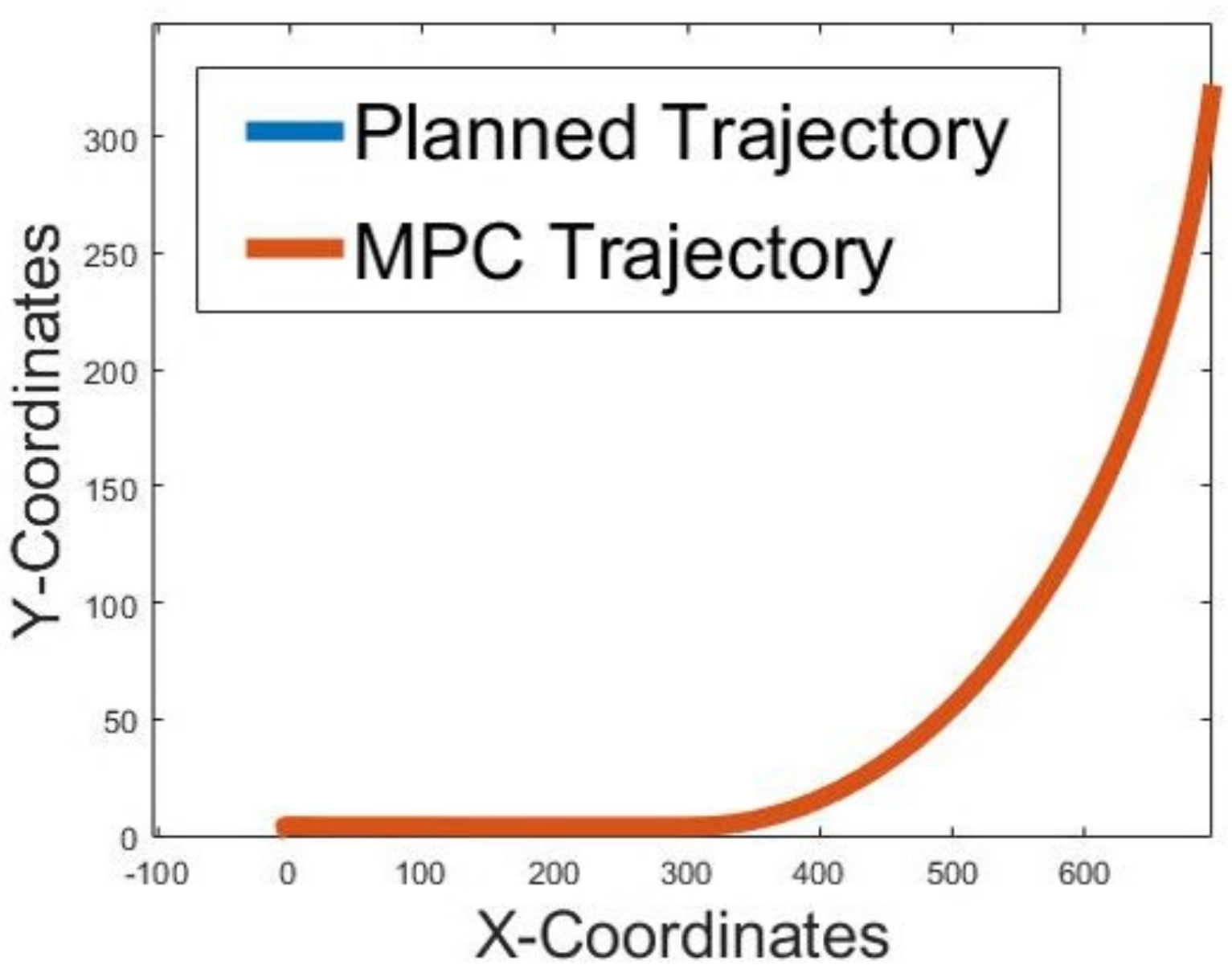}
    \, \, \, \, \, \caption{MPC baseed trajectory and planned trajectory .}
    \label{fig.}
\end{figure}

Before assessing the success of obstacle avoidance, it's essential to evaluate how well the vehicle follows the reference line when controlled by the MPC controller. Fig. 9 illustrates in the first scenario the planned trajectory alongside the actual trajectory under MPC control, with the $x-axis$ and $y-axis$ representing the real-time coordinates of the vehicle in the global coordinate system. It is evident that the two trajectories nearly coincide with each other.

After confirming that the autonomous vehicle will follow the planned route, the next step is to test the pedestrian avoidance scenario emphasized in this paper. 

\begin{figure}[!htbp]
    \centering
    \begin{subfigure}[b]{0.45\columnwidth}
        \centering
        \includegraphics[width=\textwidth]{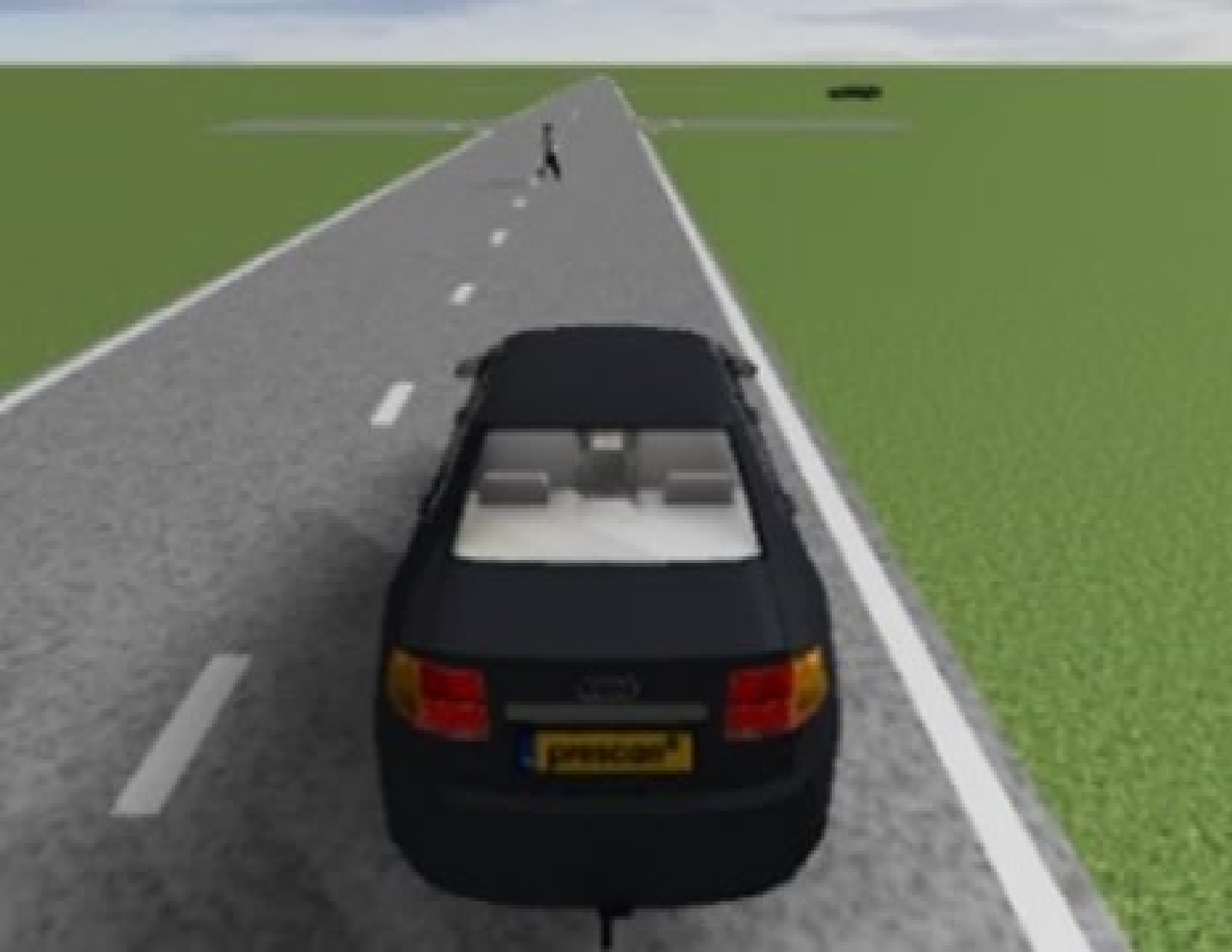}
        \caption{}
        \label{fig:1}
    \end{subfigure}
    \hfill
    \begin{subfigure}[b]{0.45\columnwidth}
        \centering
        \includegraphics[width=\textwidth]{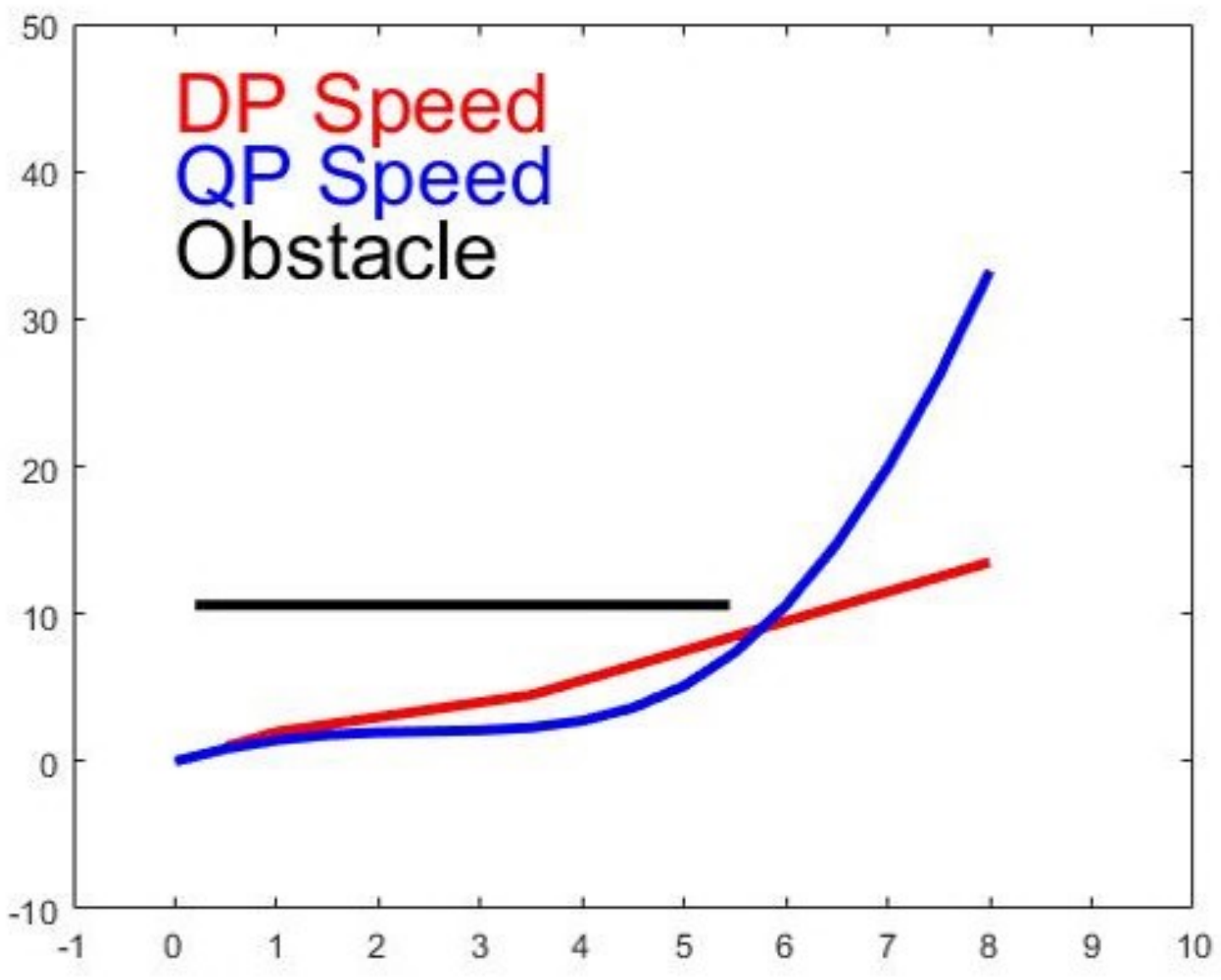}
        \caption{}
        \label{fig:2}
    \end{subfigure}

    \vspace{0.5cm} 
    \begin{subfigure}[b]{0.45\columnwidth}
        \centering
        \includegraphics[width=\textwidth]{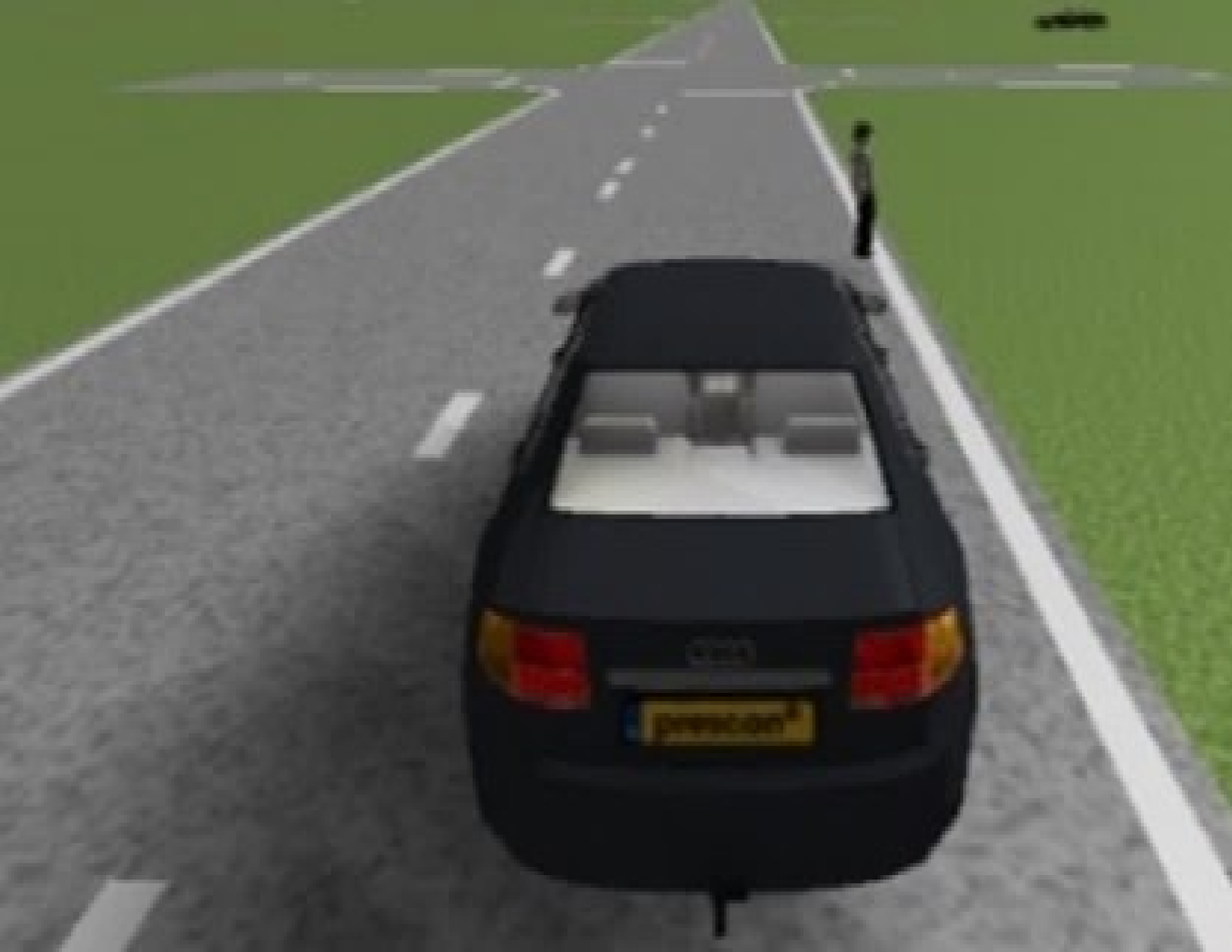}
        \caption{}
        \label{fig:3}
    \end{subfigure}
    \hfill
    \begin{subfigure}[b]{0.45\columnwidth}
        \centering
        \includegraphics[width=\textwidth]{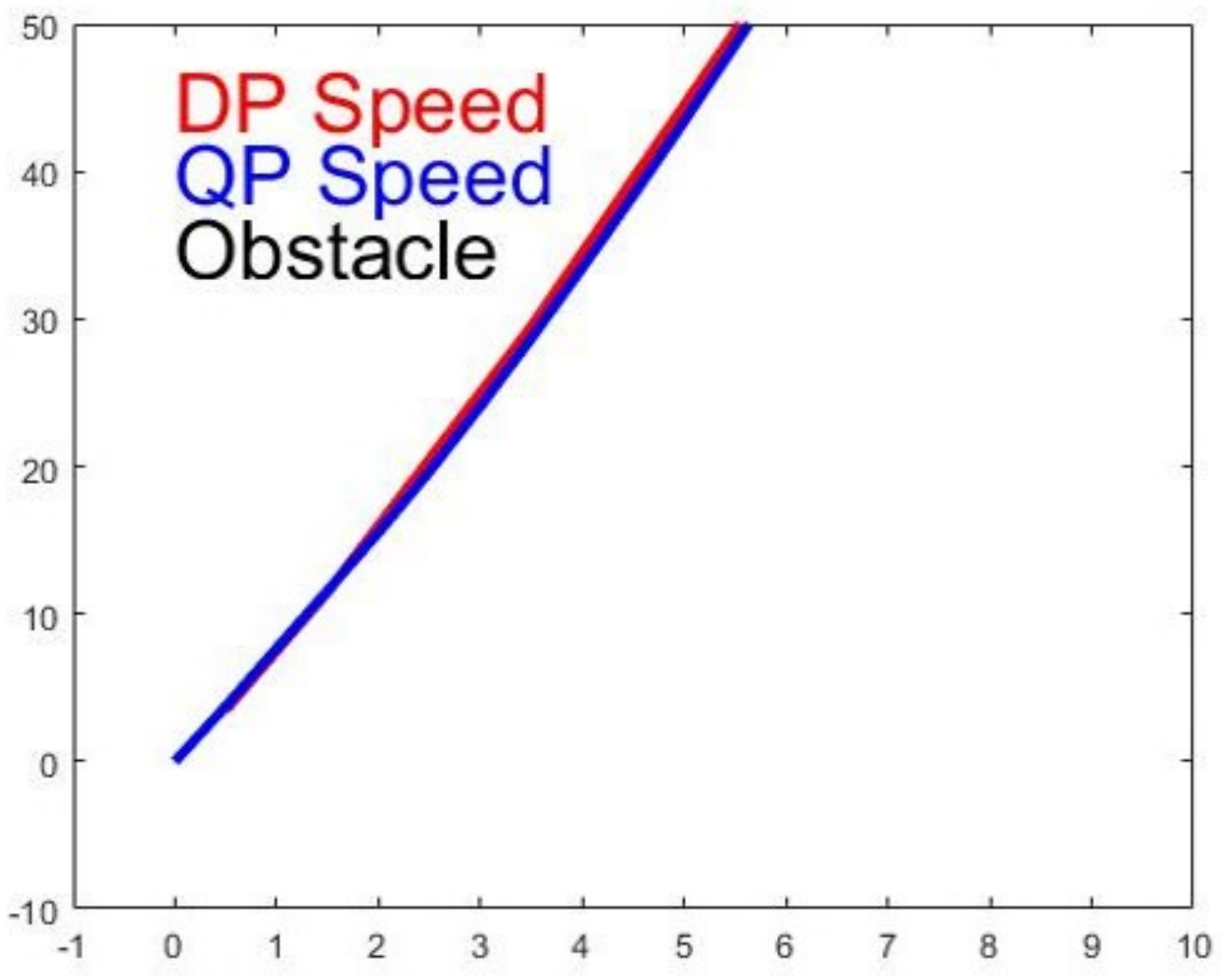}
        \caption{}
        \label{fig:4}
    \end{subfigure}

    \caption{S-T graph before and after the pedestrian.}
    \label{fig:shared}
\end{figure}

In Fig. 10(a), a pedestrian is seen crossing the road directly in front of the vehicle. Fig. 10(b) illustrates the interference range of the obstacle along with the planning intervals for both DP and QP, presented in the S-T plot. The red and blue lines indicate the speeds planned by the vehicle using DP and QP, respectively. These lines closely match the speed range of the obstacle (depicted by the black line) to avoid any potential collisions with the pedestrian. Fig. 10(c) and Fig. 10(d) illustrate that once the pedestrian has moved off the vehicle's reference line and come to a stop, only the speeds planned by DP and QP are visible on the S-T plot. The black line representing the obstacle is no longer present, signifying that the vehicle can continue on its pre-defined path and speed.

\begin{figure}[htbp]
    \centering
    \begin{subfigure}[b]{0.45\linewidth}
        \centering
        \includegraphics[width=\linewidth]{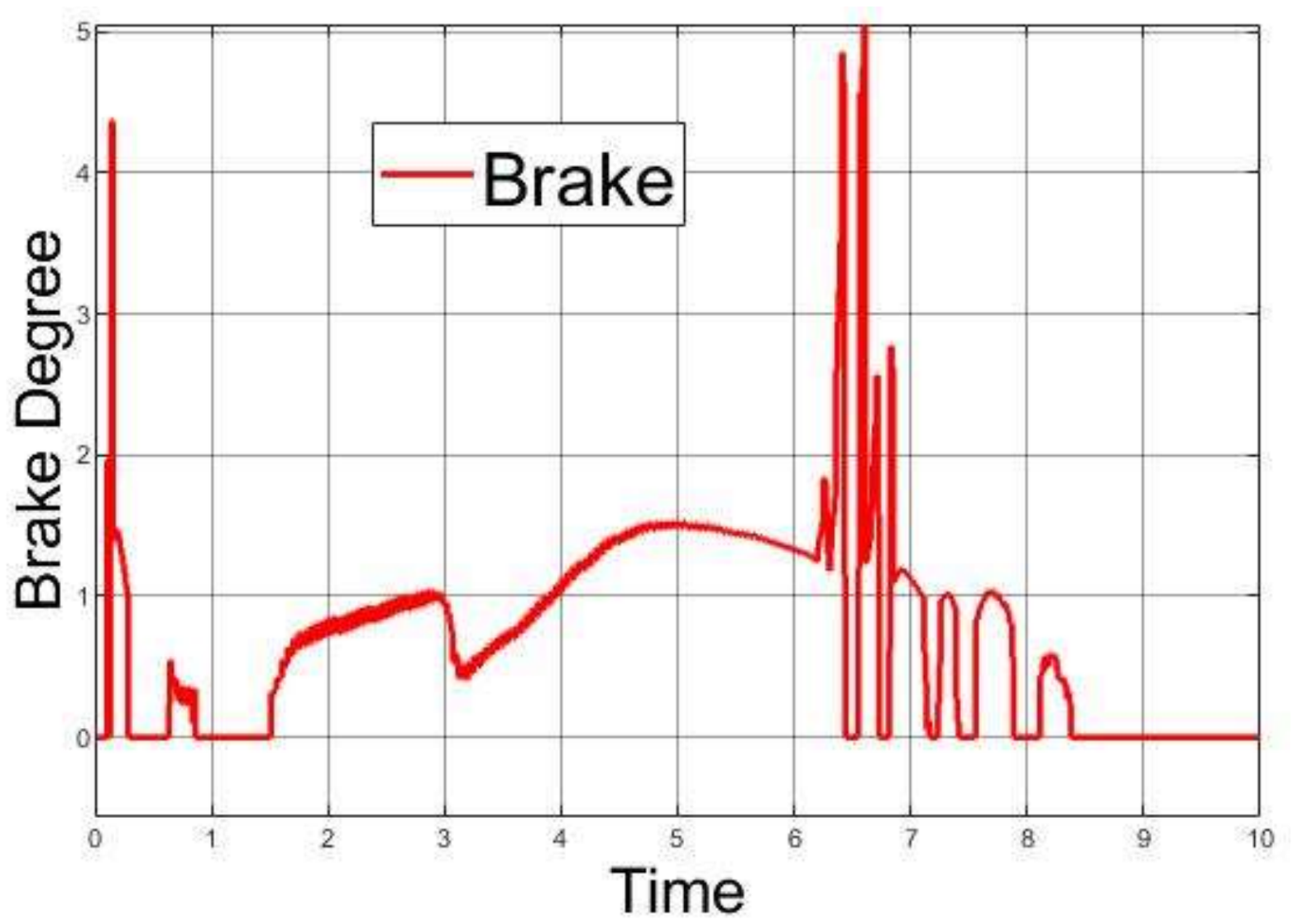} 
        \caption{}
        \label{fig:subfig1}
    \end{subfigure}
    \hfill 
    \begin{subfigure}[b]{0.45\linewidth}
        \centering
        \includegraphics[width=\linewidth]{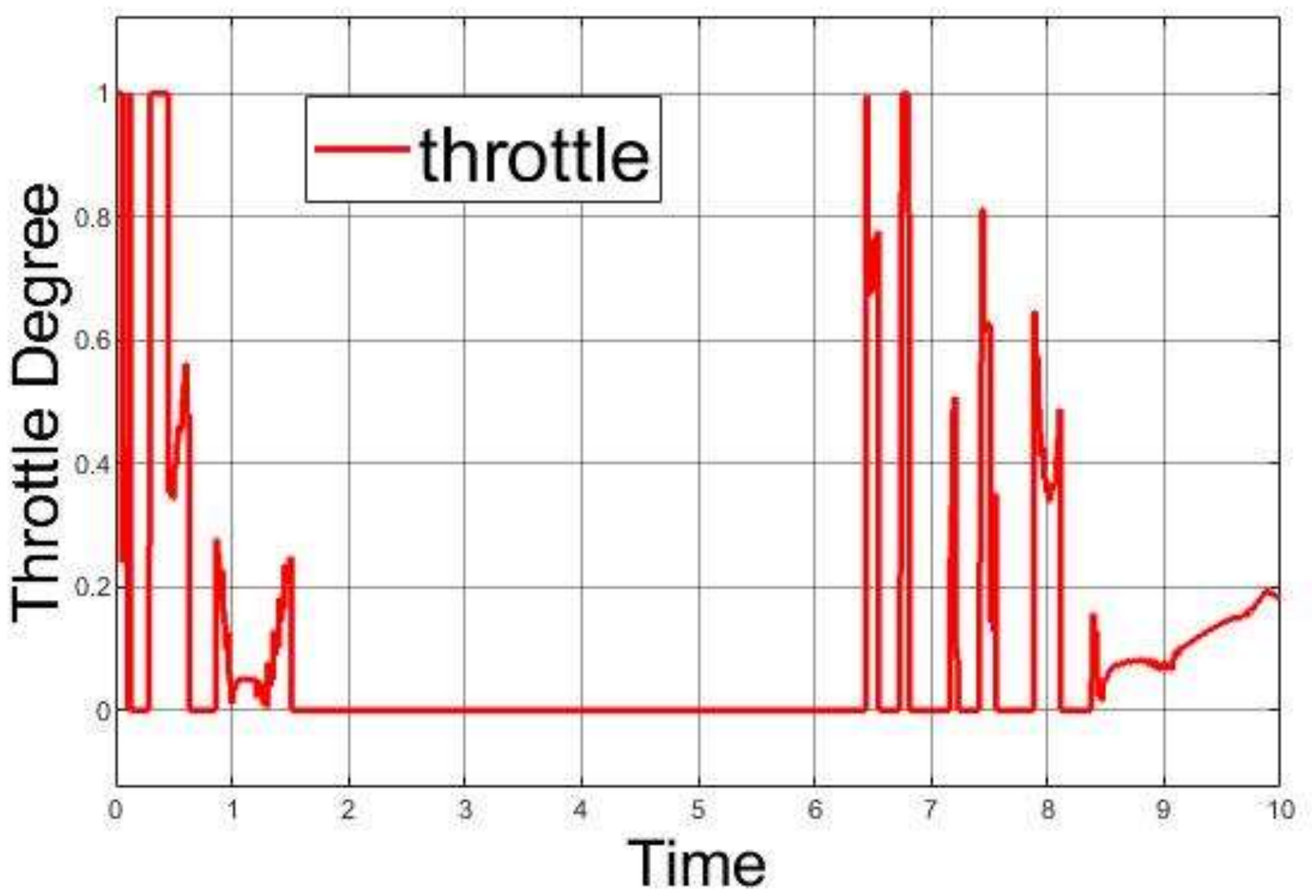} 
        \caption{}
        \label{fig:subfig2}
    \end{subfigure}
    \caption{Throttle and brake signal logs.}
    \label{fig:common}
\end{figure}

Fig. 12(a) and Fig. 12(b) capture the vehicle's deceleration to avoid a moving obstacle, specifically a pedestrian. It is evident that the vehicle consistently applies the brakes for up to 4 seconds before the pedestrian exits the road, while the throttle intermittently maintains speed. Once the pedestrian has left the roadway, the braking ceases, and the throttle remains engaged, enabling the vehicle to accelerate away.

The outcomes of the two tests indicate that the dynamics-based autonomous driving framework, which incorporates vehicle lateral and longitudinal control alongside decision-making and planning algorithms using DP and QP, effectively facilitates trajectory tracking and initial obstacle avoidance for dynamic obstacles.

\section{Conclusions}

This paper presents a comprehensive control system for autonomous vehicles that integrates both longitudinal and lateral control strategies. The proposed dual-loop PID controller effectively manages throttle and brake operations, ensuring stable longitudinal dynamics. For lateral control, the implementation of a DLQR controller enhances steering responsiveness, contributing to overall vehicle stability. Additionally, the integration of MPC controller facilitates precise trajectory tracking while adapting to dynamic environments. Finally, the use of DP and QP algorithms enables pedestrian avoidance functionality. Through two test scenarios designed in this paper, it has been demonstrated that this integrated control framework outperforms traditional control methods, offering improved safety and navigation capabilities, making it highly suitable for trajectory tracking and obstacle avoidance in autonomous driving. Future work will focus on two key directions: 1) validating the control system's performance in real-world scenarios with varying traffic conditions; and 2) exploring advanced sensor fusion techniques to further enhance situational awareness and obstacle avoidance capabilities.

\bibliographystyle{IEEEtran}      


\end{document}